\pdfoutput=1
\documentclass[11pt]{article}

\usepackage[final]{acl}

\usepackage{times}
\usepackage{latexsym}

\usepackage{booktabs}
\usepackage{multirow}
\usepackage[table]{xcolor}
\usepackage{siunitx}

\usepackage[T1]{fontenc}

\usepackage[utf8]{inputenc}

\usepackage{microtype}

\usepackage{inconsolata}

\usepackage{graphicx}

\usepackage{amsmath}
\usepackage{amssymb}

\usepackage{todonotes}

\usepackage{mdframed}

\title{
Benchmarking Debiasing Methods for LLM-based Parameter Estimates
}

\def\authorsep{\hspace{0.3em}}

\author{Nicolas Audinet de Pieuchon$^{1,4}$ \authorsep 
Adel Daoud$^{1,2}$ \authorsep
Connor T. \ Jerzak$^3$\\
{\bf Moa Johansson}$^{1,4}$ \authorsep 
{\bf Richard Johansson}$^{1,4}$ \authorsep 
\\
$^1$Chalmers University of Technology, Sweden \authorsep
$^2$Linköping University, Sweden \\
$^3$University of Texas at Austin, USA \authorsep
$^4$University of Gothenburg, Sweden\\
\tt{\{nicolas.audinet, daoud, moa.johansson, richajo\}@chalmers.se},\\
\tt{connor.jerzak@austin.utexas.edu}}

\DeclareMathOperator*{\E}{\mathbb{E}}

\newcommand{\corpus}{\mathcal{D}}

\newcommand{\numDocuments}{N}
\newcommand{\numExperts}{n}

\newcommand{\doc}{d} 

\newcommand{\X}{\mathbf{X}} 
\newcommand{\x}{\mathbf{x}} 

\newcommand{\Ygen}{\widehat{Y}} 
\newcommand{\ygen}{\hat{y}} 

\newcommand{\Yexp}{Y} 
\newcommand{\yexp}{y} 

\newcommand{\estimator}{f} 
\newcommand{\reference}{\theta^*} 
\newcommand{\classical}{\theta_\dagger} 
\newcommand{\imputation}{\tilde{\theta}} 
\newcommand{\ppi}{\mathrm{PPI}} 
\newcommand{\dsl}{\mathrm{DSL}} 

\usepackage{color,soul}

\usepackage{fancyhdr}
\fancypagestyle{firstpage}{
  \fancyhf{} 
  \fancyfoot[L]{\footnotesize To appear as: Nicolas Audinet de Pieuchon, Adel Daoud, Connor T. Jerzak, 
  Moa Johansson, Richard Johansson. 
  \emph{Proceedings of the 2025 Conference on Empirical Methods in Natural Language Processing (EMNLP), 2025.}}
}

\begin{document}

\maketitle
\thispagestyle{firstpage}

\begin{abstract}
Large language models (LLMs) offer an inexpensive yet powerful way to annotate text, but are often inconsistent when compared with experts. These errors can bias downstream estimates of population parameters such as regression coefficients and causal effects. To mitigate this bias, researchers have developed debiasing methods such as Design-based Supervised Learning (DSL) and Prediction-Powered Inference (PPI), which promise valid estimation by combining LLM annotations with a limited number of expensive expert annotations.

Although these methods produce consistent estimates under theoretical assumptions, it is unknown how they compare in finite samples of sizes encountered in applied research. We make two contributions: First, we study how each method’s performance scales with the number of expert annotations, highlighting regimes where LLM bias or limited expert labels significantly affect results. Second, we compare DSL and PPI across a range of tasks, finding that although both achieve low bias with large datasets, DSL often outperforms PPI on bias reduction and empirical efficiency, but its performance is less consistent across datasets. Our findings indicate that there is a bias-variance tradeoff at the level of debiasing methods, calling for more research on developing metrics for quantifying their efficiency in finite samples. 
\end{abstract}

\section{Introduction}\label{sec:introduction}

Large language models (LLMs) are transforming disciplines that use text as a form of evidence in testing theories, something particularly evident in computational social science \cite{ziems2024large,tornberg2024best,bail2024can,argyle2023out}. LLMs are being used to extract features critical for substantive research questions, across a myriad of domains, from measuring political ideology \cite{sim2013measuring}, style and tone of writing  \cite{el2016learning}, level of politeness \cite{priya2024computational},  the likelihood of epidemiological events  \cite{kino2021scoping}, to describing neighborhoods' health and living conditions  \cite{murugaboopathyPlatonicRepresentationsPoverty2025a}, and beyond. 
Although the use of LLMs promises to speed up the process of annotating these variables, which would previously have required time-consuming hand annotation by experts, if LLMs provide a wrong or suboptimal answer (i.e., a biased reply), downstream scientific estimates will also be biased \cite{egami2024,ppi_original}.  

Thus, although LLMs are powerful, these models often annotate in a way that is inconsistent with expert annotators \cite{pieuchon2024can,lin2025risks}. 
The distribution of LLM annotation errors can be heterogeneous or correlated with other variables of interest. 
These errors then lead to misleading substantive interpretations \cite{precisely_inaccurate}.

To handle these biases,  \emph{debiasing methods}\footnote{We stress that in this paper, the term \emph{bias} refers to an incorrectly estimated parameter in a statistical model, and a \emph{debiasing} method corrects the misestimation. We do not consider \emph{bias} in the sense of e.g. demographic biases in NLP representations.} have been developed, most prominently Prediction-Powered Inference (PPI) \cite{ppi_original} and Design-based Supervised Learning (DSL) \cite{dsl_original,egami2024}. Both frameworks produce an unbiased estimate by combining the LLM annotations with a smaller set of expert annotations. The biases in LLM-based estimates are then corrected by comparing the two sets of annotations for the subset of samples that have both the LLM (predicted) annotation and the expert's annotation. 

Debiasing methods have been shown to work in large (population) samples \cite{ppi_original,egami2024}, yet there is a lack of knowledge about \emph{when} and \emph{how much} debiasing methods provide added value in finite samples---which is what most domain researchers have at their disposal. There are no closed-form expressions that relate a debiasing method's efficacy to the allocation of expert versus model-generated annotations, leaving practitioners without analytic guidance on when one should prefer DSL or PPI over simply collecting more expert annotations. This lack of guidance could, in turn, hamper the uptake of debiasing methods or reinforce misuse of LLMs in applied scientific domains.   

Accordingly, to address this lack, we articulate the following research questions:

\vspace{0.1cm}

\noindent{\textbf{RQ1:}} When is a debiased, large‐scale LLM annotation dataset statistically preferable to a finite expert‐only dataset for unbiased estimation of a population parameter?

\noindent{\textbf{RQ2:}} What are the performance differences between the debiasing methods, and how do they vary across datasets and LLM-based annotators?

We tackle these questions by comparing PPI and DSL across four datasets and four annotation procedures. To our knowledge, ours is the first effort to compare debiasing methods empirically.
In foreshadowing our results, our analysis shows that compared to PPI, DSL achieves better debiasing results on average, but it is also the most variable in performance. Thus, PPI has a higher degree of stability; DSL is less consistent in gains. Our findings call for more research into the advantages and disadvantages of various debiasing methods across dataset and feature configurations.

\section{Background: Methods for Debiasing LLM-based Estimates}\label{sec:theory}

\begin{figure}
    \centering
    \includegraphics[width=\linewidth]{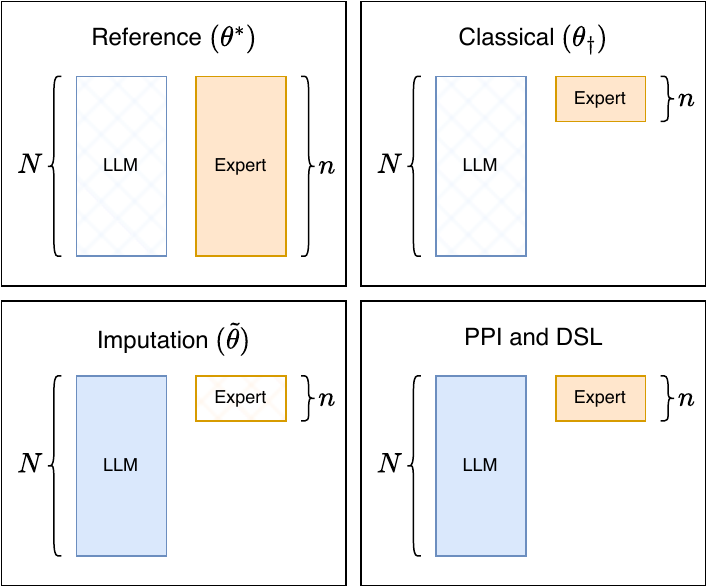}
    \caption{ 
    The reference model (top-left) is estimated from expert annotations ({\color{orange}orange}) for all $N$ samples in the dataset (i.e., $n=N$, with full expert labeling). The classical model (top-right) uses only the $n$ expert samples for downstream estimation ($n\ll N$). The imputation model (bottom-left) uses only the generated annotations for all $N$ samples ({\color{blue}blue}). The \emph{debiased} models (bottom-right) use \emph{both} LLM annotations for all $N$ samples and expert annotations for the subset of $n$ samples.
    }
    \label{fig:model_types}
\end{figure}

Let $\corpus = \{(\doc_i, \x_i, \ygen_i)\}_{i=1}^\numDocuments$ be a corpus of $\numDocuments$ documents $\doc_i$, with associated independent variables $\x_i \in \X = \{\x_i\}_{i=1}^\numDocuments$ and LLM annotations $\ygen_i \in \Ygen=\{\ygen_i\}_{i=1}^\numDocuments$. A subset of $\corpus$ of size $\numExperts$ also has additional expert annotations $\yexp_j \in \Yexp=\{\yexp_j\}_{j=1}^\numExperts$, where $\numExperts \leq \numDocuments$. Expert annotations are taken to be the ground truth and are generally costly \cite{gilardi2023chatgpt}.

Next, we focus on a general parameter of interest $\theta$, which represents the result of the downstream statistical analysis. For example, this could be a regression coefficient or a class prevalence rate. The goal of the debiasing methods is to create an estimator $\estimator$ which estimates $\theta$ based on $\X$, $\Yexp$, and $\Ygen$. Ideally, the estimator should be \textit{consistent}, meaning that $\estimator(\X, \Yexp, \Ygen) \to \theta$ as $\numDocuments \to \infty$, and \textit{precise}, meaning that we want to keep the variance and confidence intervals as small as possible.

One way to achieve this would be to ignore $\Ygen$ entirely and only use the unbiased expert annotations $\Yexp$. We call this the \textit{classical estimator} $\classical = f(\X, \Yexp)$, which is usually generated by minimizing a loss. Although this estimator produces unbiased estimates, it can have a large variance if we have few expert annotations. We call the classical estimator trained with expert annotations for \textit{all} $\numDocuments$ samples the \textit{reference estimator} $\reference$, which corresponds to the ideal but costly model that the debiasing methods are aiming towards.

Another approach would be to only use LLM annotations $\Ygen$ and ignore the expert annotations $\Yexp$. We call this the \textit{imputation estimator}, $\imputation = \estimator(\X, \Ygen)$. Here, we rely on the assumption that we can exchange the expert annotations for the LLM annotations. The hope is that, while LLM annotations might be noisier than expert annotations, we can counteract the noise by simply generating as many labels as needed, given a large enough corpus. However, the LLM may exhibit systematic biases different from those of the expert human annotators, meaning that $|\imputation - \reference| > 0$ as $\numDocuments \to \infty$, and therefore this assumption does not hold in general. 
In turn, this leads to a biased downstream estimate, and one runs the risk of being ``precisely inaccurate'' \cite{precisely_inaccurate}.

A third approach claims to be both unbiased and more precise than $\classical$. Such methods typically work by estimating parameters on LLM annotations, with a \emph{rectifier} constructed from the difference between the generated and expert annotations for the subset of the corpus for which we have both (see Figure \ref{fig:model_types}). In this paper, we investigate PPI and DSL as two of the most prominent among these methods.

\paragraph{Prediction-Powered Inference (PPI).}

PPI offers a protocol for integrating LLM predictions into downstream statistical inference via first-order debiasing \cite{ppi_original}. It begins by treating the  LLM predictions as if they were true labels and forming the ``imputation estimate'':
\(
\tilde{\theta}
\;=\;
\textrm{argmin}_{\theta}\;
\frac{1}{\numDocuments}\sum_{i=1}^{\numDocuments}
\ell_{\theta}\bigl( \x_i,\, \ygen_{i}\bigr),
\)
where $\ell_{\theta}$ is the loss defining our estimand, such as the binary cross-entropy for a logistic regression.
In general $\tilde{\theta}$ is biased, so PPI introduces the \emph{rectifier}, which, in the one parameter case equals 
\[
r_{\theta}
\;=\;
\E\bigl[
\nabla_{\theta}\ell_{\theta}(\x_i,\yexp_i)
\;-\;
\nabla_{\theta}\ell_{\theta}\bigl(\x_i,\ygen_i\bigr),
\bigr],
\]
the gradient terms capturing the systematic distortion from substituting $\ygen_i$ for the true $\yexp_i$ (the gradient difference reveals the bias direction in parameter space, which we then offset to debias). We estimate $r_{\theta}$ on the labeled sample and estimate the imputed gradient on the unlabeled set using plugin estimators. The final, first-order debiased estimate is then $\tilde{\theta} - \widehat{r}_{\theta}$. Because $\widehat r_{\theta}$ is estimated from sample averages, confidence sets can be readily obtained.

\paragraph{Design-based Supervised Learning (DSL).}

DSL \cite{dsl_original,egami2024} adopts a \emph{design‐based sampling} scheme, which assumes
\(
\pi(\ygen_i,\x_i)=\Pr(b_i=1\mid \ygen_i,\x_i) >0,
\)
where $b_i \in \{0,1\}$ denotes whether document $i$ is labeled by experts and 
where $\pi(\cdot)$ is known. The data is partitioned into $K$ folds and used to cross‐fit \(\widehat g_k\), a model to predict $\yexp_i$ as a function of $\ygen_i$ and $\x_i$:
\begin{equation*}
\widetilde y_i^k
\;=\;
\widehat g_k(\ygen_i, \x_i)
\;+\;
\frac{b_i}{\pi(\ygen_i,\x_i)}\bigl(\yexp_i-\widehat g_k(\ygen_i, \x_i)\bigr).
\label{eq:pseudo}
\end{equation*}
Then, $
\mathbb{E}\bigl[\widetilde y_i \mid \ygen_i,\x_i\bigr]
=
\mathbb{E}\bigl[\yexp_i \mid \ygen_i,\x_i\bigr]
$
regardless of misspecification of \(\widehat g_k\) via double robustness. 

Many estimands admit a moment equation form:
\(\mathbb{E}\bigl[m(\yexp_i,\x_i;\theta)\bigr]=0\) (e.g., maximum likelihood). DSL solves the empirical analogue of the moment condition with the debiased outcome, using
$\sum_{i=1}^N
m\bigl(\widetilde y_i,\x_i;\theta\bigr)
=0$, where each \(\widetilde y_i\) is constructed as above. Cross-fitting and M-estimation theory then yield consistent ``sandwich'' estimators of variance, providing valid confidence intervals.

Table~\ref{tab:ppi-dsl-contrast} summarizes the key inferential properties of PPI and DSL, highlighting similarities and differences.

\section{Methodology}
\label{sec:experiments}

\begin{figure}
\centering
\includegraphics[width=\linewidth]{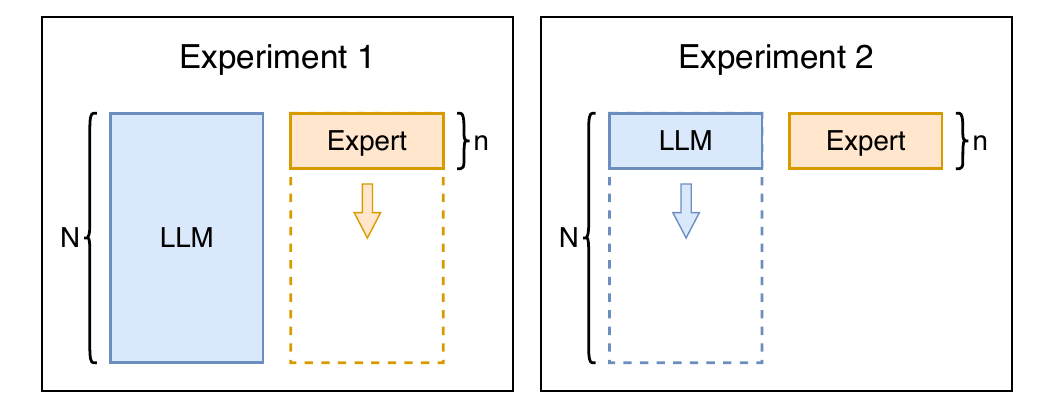}
    \caption{
    Setup for Experiments 1 and 2. Left panel (Experiment 1): Fixed $N$ total samples with LLM annotations ({\color{blue}blue}); vary $n \ll N$ expert annotations ({\color{orange}orange}) for debiasing. Right panel (Experiment 2): Vary $N$ total samples with LLM annotations ({\color{blue}blue}); fixed $n$ expert annotations ({\color{orange}orange}) for debiasing.
    }
    \label{fig:experiments}
\end{figure}

Our analysis focuses on two experiments, which we use to benchmark and contrast the $\classical$, $\ppi$, and $\dsl$ estimators (see Figure \ref{fig:experiments}). In both experiments, we focus on the coefficients of a binary logistic regression as our particular parameter of interest $\theta$. Specifically, for each dataset, we create a downstream task relating four independent variables $x_1 \ldots x_4$ to a binary outcome $y$.
The independent variables are either categorical or integers computed from text features. Each logistic regression, therefore, produces four coefficients $\beta_1 ... \beta_4$ and a $y$-intercept $\beta_0$ for a total of five parameters.
See Appendix \ref{appendix:methods-details} for package use details and a link to the code. 

\paragraph{Experiment 1.}

Our first experiment involves varying the number of expert annotations while keeping the total number of samples constant (see Figure \ref{fig:experiments}, left). Our goal here is to answer the question: how do the debiasing methods improve with an increasing proportion of expert annotations? In other words, if one has a fixed number of data samples, how much budget should one allocate towards the expert annotations for debiasing?

For this experiment, we vary the number of expert samples logarithmically. We use a minimum of 200 expert annotations (below that threshold, debiasing methods became unstable). We additionally report the proportion of expert samples $\numExperts_i/\numDocuments$ rather than the absolute number in order to compare datasets of different sizes. 
We run 250 repetitions and report $2\sigma$ confidence intervals for each entry, dataset, and annotation procedure.

\begin{figure}
    \centering
    \includegraphics[width=\linewidth]{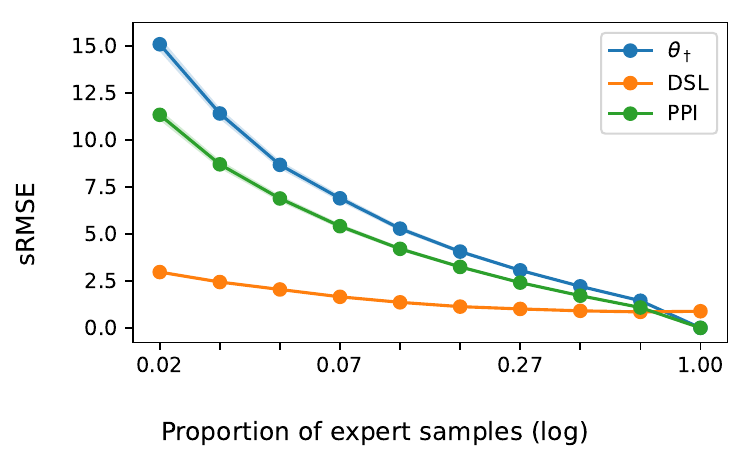}
    \caption{Results for Experiment 1 averaged over all datasets and annotation methods.}
    \label{fig:expert-all}
\end{figure}

\begin{figure*}
    \centering
    \includegraphics[width=\textwidth]{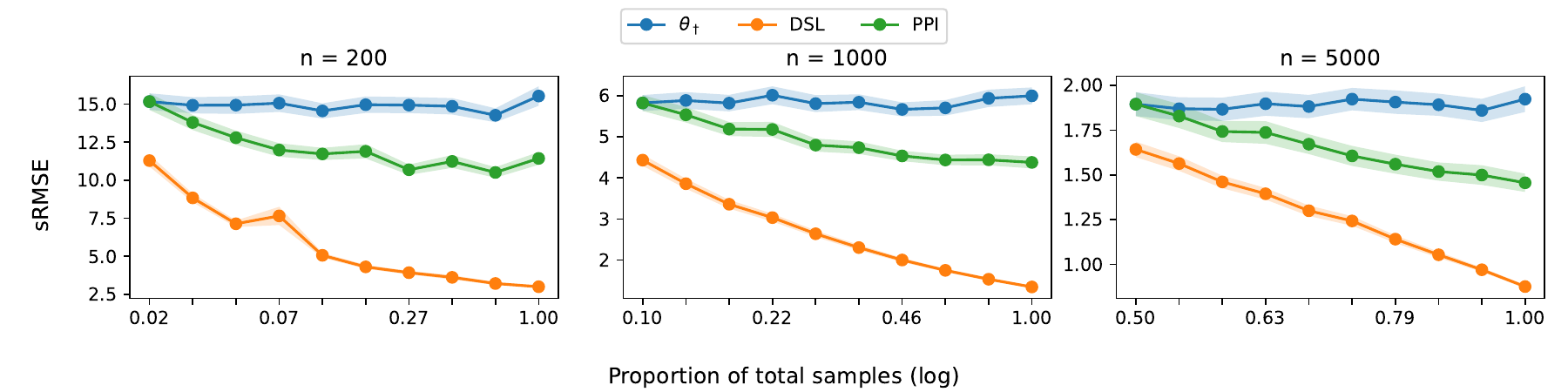}
    \caption{Results from the set of experiments varying the total number of samples, averaged over datasets and annotation methods. The $x$-axis shows the total number of samples ($N$) as a proportion of the total available samples in each dataset. The $y$-axis shows the sRMSE. The plots show results for $n=200$, $n=1000$, and $n=5000$.
}
    \label{fig:total-all}
\end{figure*}

\paragraph{Experiment 2.}

In our second experiment, we vary the total number of samples while keeping the number of expert annotations fixed (see Figure \ref{fig:experiments}, right). This setup targets scenarios where the expert annotation budget is limited but unlabeled data is abundant---such as in large news or social-media corpora---enabling practitioners to evaluate how additional unlabeled data enhances the effective sample size (i.e., the equivalent number of expert annotations needed to match the debiased estimator's precision) via debiasing methods. Specifically, we ask: given a fixed expert budget, how much does the effective sample size increase with more generated annotations? We repeat these experiments using 200, 1,000, and 5,000 expert annotations.

Like in Experiment 1, we vary the total number of samples logarithmically. The minimum number of total samples is defined by the number of available expert samples. The maximum number of total samples is determined by the size of the available dataset, which varies. We report the proportion of total samples with respect to the total number of available samples to facilitate comparison between datasets. We use 250 repetitions to estimate the $2\sigma$ confidence interval.

\paragraph{Datasets and Annotations.}

We replicate our experiments over four datasets: Multi-domain Sentiment, Misinfo-general, Bias in Biographies, and Germeval18 (see Appendix~\ref{appendix:dataset-description}). We also compare performance across four LLM-model classes: BERT, DeepSeek v3, Phi-4, and Claude 3.7 Sonnet (see Appendix \ref{appendix:model-details}). Input variables are either additional annotations available from the original dataset or quantities derived from the text, such as the text length in characters. We compare $\ppi$, $\dsl$, and $\classical$ with the same number of annotations. The datasets are available at \href{https://huggingface.co/datasets/nicaudinet/llm-debiasing-benchmark}{\url{https://huggingface.co/datasets/nicaudinet/llm-debiasing-benchmark}}. 

\paragraph{Evaluation Metrics.}

We evaluate performance of debiasing methods by comparing the respective models against the reference model $\reference$. Comparison between models is done using a standardized Root Mean Squared Error (sRMSE), which captures both bias and variance for a holistic performance assessment (see Appendix \ref{s:EvalMetricsDetails}). We standardize by scaling according to the reference model coefficients. 

\section{Results}\label{s:Results}

In our experiments, we contrasted $\classical$, PPI, and DSL with the reference model $\reference$.
The only difference between $\classical$ and $\reference$ is that they are trained on a different number of expert annotations --- $\classical$ is trained on only the expert annotations that would have been given to one of the debiasing methods. Accordingly, the smaller the proportion of expert annotations given to the debiasing methods, the more inaccurate $\classical$ becomes, which is reflected as a high sRMSE. As we increase the proportion of expert annotations, $\classical$ converges towards $\reference$, and we observe a monotonically decreasing sRMSE. At a proportion of 1, there is no difference between $\classical$ and $\reference$ (the sRMSE is 0).

Results of Experiment 1 are displayed in Figure \ref{fig:expert-all}. We observe that PPI has a lower sRMSE than $\classical$ for all data points. This is expected and, under assumptions, guaranteed by theory. DSL exhibits a significantly lower sRMSE than both PPI and $\classical$ for almost all data points, showing that it is able to use the expert annotations more efficiently than both. However, the crossing at the end, when virtually all expert annotations contribute to the debiasing procedure, is curious: why do the DSL and $\classical$ curves cross? Further analysis (see Appendix \ref{sec:AppendixII}) reveals that this crossing phenomenon is dataset-dependent. In particular, for the Misinfo-general dataset, DSL performs worse than both PPI and $\classical$ for all samples.

A complete explanation of this phenomenon is still unknown. We have ruled out hypotheses related to preprocessing (e.g., centering); we have not identified obvious properties of the dataset that predict anomalous DSL estimates (e.g., agreement between expert and LLM annotations). Emerging evidence, however, points to multicollinearity in the feature set as a contributing factor: DSL appears more sensitive to it than PPI, which is comparatively robust. In particular, we find in Figure \ref{fig:expert-all-no-collinear} in Appendix \ref{sec:removing-collinear-variables} that the detrimental scaling of DSL in the Misinfo-general dataset is greatly improved when we remove highly collinear features. We also find in Figure \ref{fig:bias-all-no-collinear} in Appendix \ref{appendix:bias-plots} that bias decreases for all three methods when removing highly collinear features. Another remaining explanation is that, although PPI debiasing via subgradients leverages less information compared to DSL (which uses external sampling design knowledge), it avoids instabilities commonly associated with weighting estimators \citep{zubizarreta2015stable}. Future work should explore these and related explanations, including how feature correlations interact with debiasing stability.


The results of Experiment 2 are displayed in Figure \ref{fig:total-all}. Since $\classical$ does not use generated annotations, its sRMSE remains constant as the dataset size grows. We also observe that PPI and DSL both outperform $\classical$ in each of the three cases; performance of both tends to improve as we increase the total dataset size. 

To translate these empirical findings into practical guidance for resource allocation in computational social science \cite{daoud_statistical_2023}, we adapt the budgeting template from \citet{broska2025mixed}'s mixed subjects design, which optimizes the mix of costly expert annotations and cheaper LLM predictions based on their correlation and error profiles. Expert labeling on Amazon Mechanical Turk currently averages \$0.10 per label as of 2025 \citep{admon2025_data-annotation-costs}. For LLM inference on our largest corpus (Bias in Biographies, $N=10{,}000$), assuming 300 input tokens per document (3 million total input tokens) and 10 output tokens per prompt (0.1 million total output tokens): Phi-4 incurs \$0.06 per million input tokens and \$0.14 per million output tokens, yielding a total cost of $\approx$ \$0.20 (equivalent to 2 expert labels). DeepSeek v3, at \$0.56 per million input tokens and \$1.68 per million output tokens, costs $\approx$ \$2 (20 expert labels). Claude 3.7 Sonnet, at \$3 per million input tokens and \$15 per million output tokens, costs $\approx$ \$10.50 (105 expert labels). BERT fine-tuning adds negligible cloud costs ($\approx$ \$0.50 USD, 5 expert labels), with debiasing computations (DSL/PPI) under \$1 total on standard hardware. In this 10,000-document scenario, the break-even $n$ for cost (where $n$ expert labels cost as much as full-model inference) is thus 2 for Phi-4, 20 for DeepSeek, 105 for Claude, and 5 for BERT---far below full expert annotation. 
Given our results (e.g., sRMSE $<0.2$ at $n=200$), we encourage practitioners to supplement their analyses with LLM predictions starting at these thresholds.

\section{Conclusion}

This study has investigated the performance of two LLM debiasing methods. On average, both debiasing methods produce downstream estimates that are closer to a reference model than those obtained by using a small number of expert annotations. We also observe that DSL seems to significantly outperform PPI across datasets and annotation methods. However, DSL performance appears more inconsistent and dataset-dependent. Both DSL and PPI are more efficient than relying solely on a small, human-annotated dataset, so we encourage researchers to integrate debiasing methods into their analyses for improved estimation. While DSL outperforms PPI on most datasets, its performance is more inconsistent across them; therefore, we recommend reporting results from both methods until DSL's variability is better understood.

\section*{Limitations}

Our study focuses on two specific debiasing methods, DSL and PPI, leaving out several other emerging techniques such as the recently proposed predict-then-debias (e.g., \citealp{kluger2025prediction}) and prediction-powered inference with inverse probability weighting \citep{datta2025prediction}. We only consider scenarios where the outcome variable requires annotation, thereby restricting the scope to single-task classification; we focus on binary outcomes as a simplifying assumption to facilitate benchmarking of the debiasing methods, though future work should extend this to multi-class or continuous outcomes. Future work should also consider situations where input variables are LLM-annotated or there is information leakage among variables \cite{tleakage2022}. In addition, while DSL and PPI can be applied to any M-estimator, our experimental evaluation of downstream tasks is currently limited to logistic regression (corresponding to the type of annotation we have considered). Future work should consider a variety of other statistical estimators, such as survival or hierarchical models. 

Moreover, our experiments also concentrate on four datasets with relatively short texts in English or German, so further evaluation is needed in other languages, domains, and text lengths. Lastly, we assume expert-labeled data to be the ground truth; in practice, human annotations can also be noisy or inconsistent \cite{artstein2008survey}. Future work should examine how to extend or adapt methods such as DSL and PPI when the expert labels themselves may be subject to significant measurement error or domain shifts. We also acknowledge that the robustness of these debiasing methods under worst-case or adversarial settings remains an open problem.

\section*{Acknowledgments}

We thank Naoki Egami, Katherine Keith, Brandon Stewart, and anonymous reviewers for helpful comments. This research was supported by the project \emph{Countering Bias in AI Methods in the Social Sciences} under the Wallenberg AI, Autonomous Systems and Software Program -- Humanity and Society (WASP-HS), funded by the Marianne and Marcus Wallenberg Foundation and the Marcus and Amalia Wallenberg Foundation. We acknowledge support from Alvis and Vera compute systems, provided by the National Academic Infrastructure for Supercomputing in Sweden (NAISS).

\bibliography{custom}

\appendix

\section{Datasets Description}
\label{appendix:dataset-description}

We here present information about the datasets used in the analysis. All datasets are constructed by extracting a balanced subset of publicly available datasets. Following the simulation experiment from Egami et al. \cite{dsl_original} we created four features ($x1, x2, x3, x4$) used in the downstream task to predict the annotated output $y$. Details of the original datasets and feature creation are reported below. All datasets and LLM annotations used in the paper are available at  \href{https://huggingface.co/datasets/nicaudinet/llm-debiasing-benchmark}{\url{https://huggingface.co/datasets/nicaudinet/llm-debiasing-benchmark}}.

\paragraph{Multi-domain Sentiment.} The Multi-domain Sentiment dataset is a corpus of product reviews taken from Amazon \cite{amazon_dataset}. The dataset was originally used to investigate domain adaptation in sentiment classifiers. We used a subset taken from 6 domains, consisting of 11,914 reviews with two sets of annotations: a binary sentiment label (positive or negative) and a domain label (books, camera, DVD, health, music, or software). The dataset is balanced both in sentiment and topic labels.

For the downstream task, we use the sentiment label as the outcome variable. The independent variables are: the domain label (transformed to numeric values 0-5), the number of characters in the review, the number of space-separated words in the review, and the number of repetitions of the word ``I'' in the review.

\paragraph{Misinfo-general.} The Misinfo-general dataset is a large corpus of British newspaper articles \cite{misinfo_dataset} originally used to benchmark out-of-distribution performance of misinformation models. For our experiments, we selected articles from 2022 that were published in one of two venues: The Guardian UK or The Sun. We then balanced the dataset to have 5000 articles in each class.

For the downstream task, we use the venue as the binary outcome variable. The independent variables are: the number of characters in the article, the number of space-separated words in the article, the number of capital letters in the article, and the number of characters in the title of the article.

\paragraph{Bias in Biographies.} 

The Bias in Biographies dataset is a corpus of short biographies originally used to study gender bias in occupational classification \cite{biobias_dataset}. The corpus consists of English-language online biographies from the Common Crawl, annotated with self-identified binary gender and occupation labels (with 28 categories), enabling analysis of implicit gender biases in textual representations. Here, $N=$ 10,000.

For the downstream task, we use the gender label as the outcome variable. This variable is balanced. Independent variables are: the occupation label (transformed to a numeric value, 0-27), the number of characters in the biography, the number of space-separated words in the biography, and the number of capital letters in the biography.

\paragraph{Germeval18.} 

The Germeval18 dataset is a corpus of German tweets. It was used in the GermEval shared task on the identification of offensive language in 2018 \cite{germeval18_dataset}. It is composed of a training and test set of documents with associated toxicity labels, totaling 5676 documents. We use a balanced subset of the data.

For the downstream task, we use the binary toxicity label as the outcome variable. The independent variables are: the number of characters in the tweet, the number of space-separated words in the tweet, the number of capital letters in the tweet, and the number of ``@'' characters in the tweet.

\section{Details of Evaluation Metrics}\label{s:EvalMetricsDetails}

We define the standardized Root Mean Squared Error ($sRMSE$) as:
\[
\mathrm{sRMSE}(\theta; d) = \sqrt{\mathbb{E} \left[ \left( \frac{\theta - \reference_d}{\reference_d} \right)^2 \right]}.
\]
where $\theta$ are the coefficients from the model under test and $\reference_d$ are the coefficients from the reference model for dataset $d$. 

\section{Model Details}
\label{appendix:model-details}

\paragraph{BERT + Logistic Regression.}  

As a representative of supervised approaches, we fine‐tune a pre‐trained BERT encoder \cite{devlin2019bert} on the expert‐labeled subset to obtain contextual representations $\mathbf{h}_i = \mathrm{BERT}(\doc_i)$, which are then passed to a logistic regression head trained to predict $y_i$.  

\paragraph{Large Language Models.} 

We also generate annotations with three language models: Microsoft Phi-4 \citep{abdin2024phi}, DeepSeek v3 \citep{liu2024deepseek}, and Claude 3.7 Sonnet \citep{claude-3.7-sonnet}. Phi-4 is a 14B open-weight model, which we ran locally with the default temperature of 1.0. We used the paid DeepSeek and Anthropic APIs to access DeepSeek v3 and Claude 3.7 Sonnet, respectively. We paid approximately \$10 for the DeepSeek API and approximately \$100 for the Anthropic API (2025 USD). We used the default decoding mechanism and temperature of 1.0 for both models. The prompts used to generate the labels are available in Appendix \ref{appendix:prompts}. In some cases, the annotations generated for a small number of the documents did not fit the annotation schema. These samples were ignored.

\section{Package and Code Details}
\label{appendix:methods-details}

For the classical logistic regression, we use the \texttt{scikit-learn} Python package. We use no regularization and set the maximum iterations to 1000.

For DSL, we use the \texttt{dsl} R package developed by the original paper authors for both experiments. We leave the parameters to their default settings.

For PPI, we use the \texttt{ppi\_py} Python package \cite{angelopoulos2023ppiplus}---an implementation of the \texttt{PPI+} framework by the original PPI authors---for both experiments. We also leave the parameters to their default settings.

The source code for the experiments is available at \href{https://github.com/nicaudinet/llm-debiasing-benchmark}{\url{https://github.com/nicaudinet/llm-debiasing-benchmark}}.

\section{Prompts}
\label{appendix:prompts}

Figures \ref{fig:amazon-prompt}, \ref{fig:misinfo-prompt}, \ref{fig:biobias-prompt}, and \ref{fig:germeval-prompt} show the prompt templates used to make prompts for LLM annotation. The prompt templates were specialized for each dataset since each dataset corresponds to a different annotation task. However, the structure of the prompt templates was kept the same: first, a short description of the task, then an explanation of the formatting with two simple examples, and finally the document to classify. For each dataset, we also include a system prompt (see Table \ref{tab:system-prompts}).

\begin{figure*}
\centering
\begin{mdframed}
{\small
\begin{verbatim}
Classify the following review as either:
- POSITIVE if the review indicates an overall positive sentiment
- NEGATIVE if the review indicates an overall negative sentiment

Give no other explanation for your classification, only output the label.

Here are two examples of the formatting I would like you to use, where
< REVIEW_TEXT > is a stand-in for the article text:

< REVIEW_TEXT >

CLASSIFICATION: POSITIVE

< REVIEW_TEXT >

CLASSIFICATION: NEGATIVE

Here's the review to classify:

{text}

CLASSIFICATION: 
\end{verbatim}
}
\end{mdframed}
\caption{The prompt template used to annotate documents from the Multi-domain Sentiment dataset, where \texttt{\{text\}} is substituted with the document in question.}
\label{fig:amazon-prompt}
\end{figure*}

\begin{figure*}
\begin{mdframed}
{\small
\begin{verbatim}
Classify the following article as either:
- THESUN if it is likely to have been published in the British tabloid newspaper
  The Sun
- THEGUARDIAN if it is likely to have been published in the British daily
  newspaper The Guardian

Give no other explanation for your classification, only output the label.

Here are two examples of the formatting I would like you use, where
< ARTICLE_TEXT > is a stand-in for the article text:

< ARTICLE_TEXT >

CLASSIFICATION: THESUN

< ARTICLE_TEXT >

CLASSIFICATION: THEGUARDIAN

Here's the article I would like you to classify:

{text}

CLASSIFICATION: 
\end{verbatim}
}
\end{mdframed}
\caption{The prompt template used to annotate documents from the Misinfo-general dataset, where \texttt{\{text\}} is substituted for the document in question}
\label{fig:misinfo-prompt}
\end{figure*}

\begin{figure*}
\begin{mdframed}
{\small
\begin{verbatim}
Classify the following textual biographies as either:
- MALE if the subject is likely to be male
- FEMALE if the subject is likely to be female

Give no other explanation for your classification, only output the label.

Here are two examples of the formatting I would like you use, where < BIOGRAPHY_TEXT >
is a stand-in for the textual biography:

< BIOGRAPHY_TEXT >

CLASSIFICATION: MALE

< BIOGRAPHY_TEXT >

CLASSIFICATION: FEMALE

Here's the textual biography I would like you to classify:

{text}

CLASSIFICATION: 
\end{verbatim}
}
\end{mdframed}
\caption{The prompt template used to annotate documents from the Bias in Biographies dataset, where \texttt{\{text\}} is substituted for the document in question}
\label{fig:biobias-prompt}
\end{figure*}

\begin{figure*}
\begin{mdframed}
{\small
\begin{verbatim}
Classify the following German tweets as either:
- OFFENSIVE if the tweet is likely to contain an offense or be offensive
- OTHER if the tweet is _not_ likely to contain an offense or be offensive

Give no other explanation for your classification, only output the label.

Here are two examples of the formatting I would like you use, where < TWEET_TEXT >
is a stand-in for the text of the tweet:

< TWEET_TEXT >

CLASSIFICATION: OFFENSIVE

< TWEET_TEXT >

CLASSIFICATION: OTHER

{make_examples(examples)}

Here's the German tweet I would like you to classify:

{text}

CLASSIFICATION: 
\end{verbatim}
}
\end{mdframed}
\caption{The prompt template used to annotate documents from the Germeval18 dataset, where \texttt{\{text\}} is substituted for the document in question}
\label{fig:germeval-prompt}
\end{figure*}

\begin{table*}[ht]
\centering
\begin{tabular}{l|l}
\textbf{Dataset} & \textbf{System Prompt} \\
\hline
Multi-domain Sentiment & ``You are a perfect sentiment classification system'' \\
Misinfo-general & ``You are a perfect newspaper article classification system'' \\
Bias in Biographies & ``You are a perfect biography classification system'' \\
Germeval18 & ``You are a perfect German tweet classification system'' \\
\end{tabular}
\caption{The system prompts used to annotate the various datasets}
\label{tab:system-prompts}
\end{table*}

\section{Results by Dataset}\label{sec:AppendixII}

\begin{figure*}[h]
    \centering
    \includegraphics[width=0.8\textwidth]{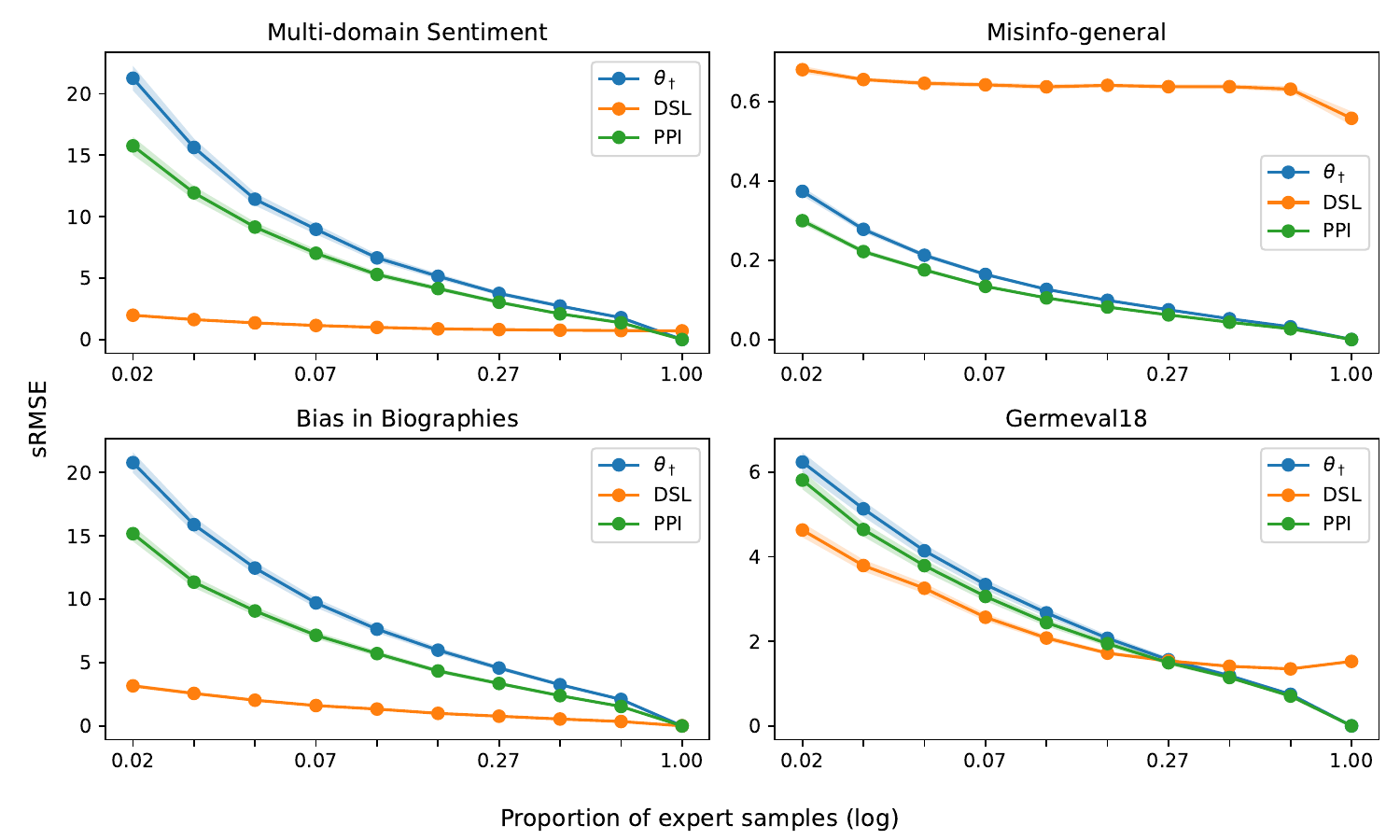}
    \caption{Results from the set of experiments varying the proportion of expert samples, aggregated per dataset.}
    \label{fig:expert-dataset}
\end{figure*}

Figure \ref{fig:expert-dataset} showcases the results for Experiment 1 broken down by dataset. In all four datasets, PPI outperforms $\classical$ for all data points. DSL outperforms PPI and $\classical$ in most cases. However, 3 of the datasets exhibit cross-over behavior for higher proportions of expert samples, with Misinfo-general being the outlier where DSL performs significantly worse than both PPI and $\classical$ for all data points.

\section{Results Removing Collinear Variables}\label{sec:removing-collinear-variables}

Here we investigated the dependence of $\dsl$ on correlations between variables. Some of the features we chose for the datasets were highly collinear (e.g.), the number of characters and the number of space-separated words in a piece of text). We gather the Pearson $r^2$ correlations in Table \ref{tab:correlations}. For each dataset and annotation type, we proceeded to remove collinear features by finding feature pairs with $r^2$ above 0.9 and removing the latter variable (for instance, we remove $x3$ for the Multi-domain Sentiment dataset).

\begin{figure*}[h]
    \centering
    \includegraphics[width=0.7\textwidth]{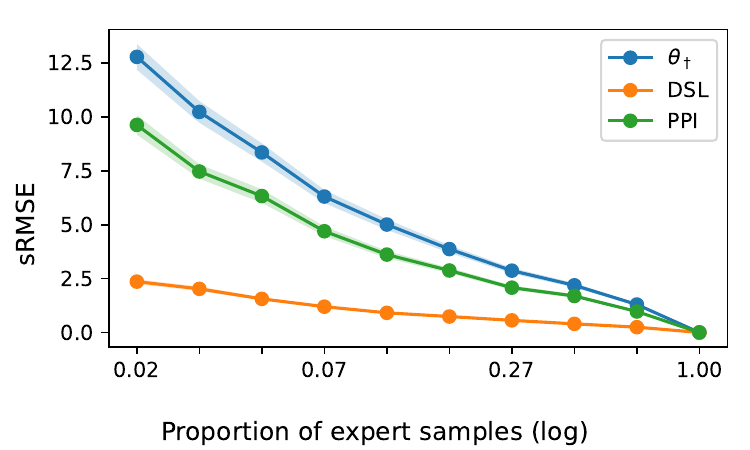}
    \caption{Performance of debiasing methods in Experiment 1 after removing highly collinear features ($r^2 > 0.9$) averaged over all datasets.}
    \label{fig:expert-all-no-collinear}
\end{figure*}

\begin{figure*}[h]
    \centering
    \includegraphics[width=\textwidth]{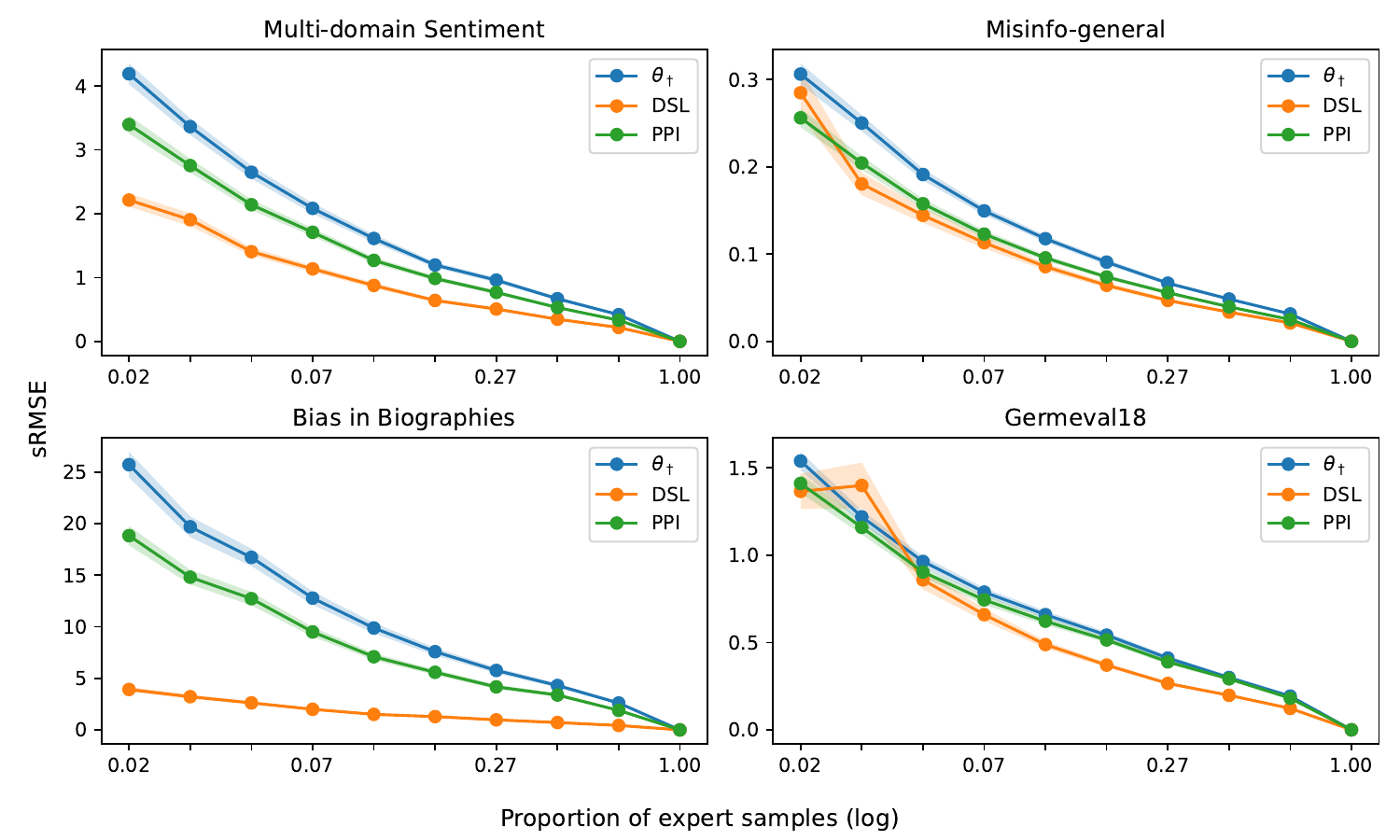}
    \caption{Performance of debiasing methods in Experiment 1 after removing highly collinear features ($r^2 > 0.9$) for each dataset.}
    \label{fig:expert-dataset-no-collinear}
\end{figure*}

The results of running Experiment 1 with the reduced datasets are shown in Figure \ref{fig:expert-all-no-collinear} and Figure \ref{fig:expert-dataset-no-collinear}. The results show that removing the collinear features mitigated the cross-over effect observed with $\dsl$ for higher proportions of expert samples.

\begin{table*}[!ht]
    \centering
    \begin{tabular}{llS[table-format=1.3]S[table-format=1.3]S[table-format=1.3]S[table-format=1.3]S[table-format=1.3]S[table-format=1.3]}
    \toprule
    Dataset & LLM & {x1,x2} & {x1,x3} & {x1,x4} & {x2,x3} & {x2,x4} & {x3,x4} \\
    \midrule
    \multirow{4}{*}{\cellcolor{white}Multi-domain Sentiment}
      & bert     & 0.008 & 0.007 & 0.013 & \textbf{0.995} & 0.253 & 0.283 \\
      & deepseek & 0.008 & 0.007 & 0.013 & \textbf{0.995} & 0.255 & 0.284 \\
      & phi4     & 0.007 & 0.007 & 0.012 & \textbf{0.995} & 0.253 & 0.283 \\
      & claude   & 0.007 & 0.007 & 0.012 & \textbf{0.995} & 0.252 & 0.282 \\
    \midrule
    \multirow{4}{*}{Misinfo-general}
      & bert     & \textbf{0.995} & 0.617 & 0.026 & 0.618 & 0.021 & 0.002 \\
      & deepseek & \textbf{0.995} & 0.621 & 0.025 & 0.622 & 0.021 & 0.002 \\
      & phi4     & \textbf{0.995} & 0.617 & 0.026 & 0.618 & 0.021 & 0.002 \\
      & claude   & \textbf{0.995} & 0.617 & 0.026 & 0.618 & 0.022 & 0.002 \\
    \midrule
    \multirow{4}{*}{Bias in Biographies}
      & bert     & 0.000 & 0.000 & 0.001 & \textbf{0.965} & 0.351 & 0.329 \\
      & deepseek & 0.000 & 0.000 & 0.001 & \textbf{0.964} & 0.346 & 0.325 \\
      & phi4     & 0.000 & 0.000 & 0.001 & \textbf{0.965} & 0.351 & 0.329 \\
      & claude   & 0.000 & 0.000 & 0.001 & \textbf{0.965} & 0.351 & 0.329 \\
    \midrule
    \multirow{4}{*}{Germeval18}
      & bert     & 0.349 & 0.250 & 0.190 & \textbf{0.961} & 0.685 & 0.653 \\
      & deepseek & 0.282 & 0.161 & 0.096 & \textbf{0.940} & 0.470 & 0.452 \\
      & phi4     & 0.332 & 0.223 & 0.161 & \textbf{0.956} & 0.615 & 0.590 \\
      & claude   & 0.284 & 0.164 & 0.101 & \textbf{0.941} & 0.487 & 0.468 \\
    \bottomrule
    \end{tabular}
    \caption{The Pearson $r^2$ correlations between each pair of features for each dataset and LLM annotations. Pairs of features with correlations above the threshold are highlighted. Correlations for the same dataset may differ slightly between LLM annotations because the LLMs failed to annotate a small portion of the samples, which we discarded.}
    \label{tab:correlations}
\end{table*}

\section{Standardized Bias Plots} 
\label{appendix:bias-plots}

We report the performance of the debiasing methods for Experiment 1 in terms of the standardized bias, following the simulation experiment from Egami et al. \citep{dsl_original}. The standardized bias is defined similarly to the sRMSE as:
$$
\text{Standardized Bias}(\theta; d) = \mathbb{E} \left[ \frac{\theta - \reference_d}{\reference_d} \right]
$$
where $\theta$ are the coefficients from the downstream task and $\reference_d$ are the coefficients from the reference model for dataset $d$.

\begin{figure*}
    \centering
    \includegraphics[width=0.7\textwidth]{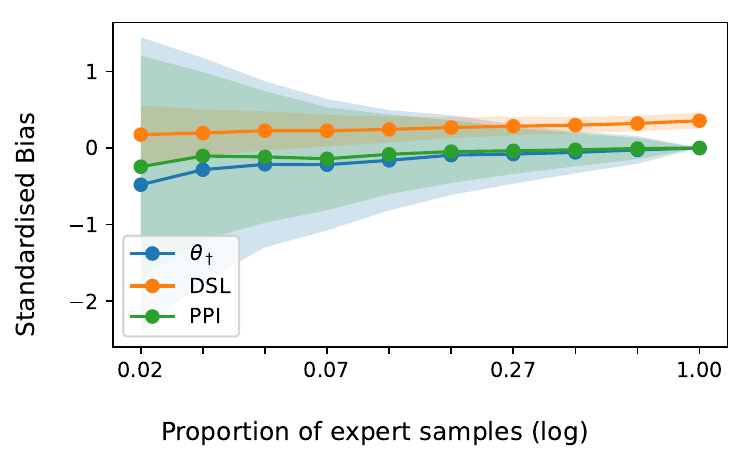}
    \caption{Performance of debiasing methods for Experiment 1 in terms of the standardized bias aggregated over all datasets.}
    \label{fig:bias-all-original}
\end{figure*}

\begin{figure*}
    \centering
    \includegraphics[width=1\textwidth]{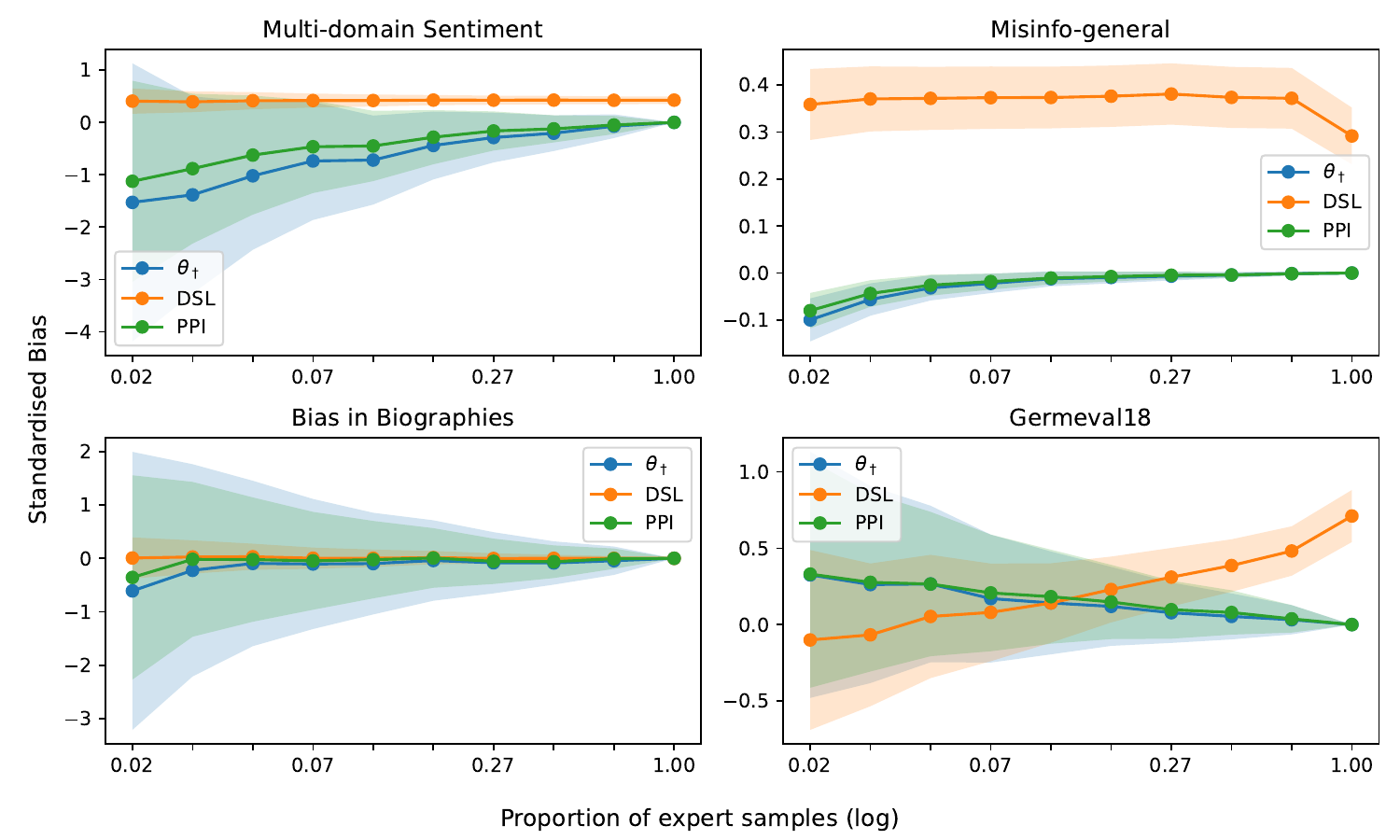}
    \caption{Performance of debiasing methods for Experiment 1 in terms of the standardized bias for each dataset.}
    \label{fig:bias-datasets-original}
\end{figure*}

Figure \ref{fig:bias-all-original} and Figure \ref{fig:bias-datasets-original} show the results for the original experiment with four features. We notice that PPI consistently produces slightly less biased coefficients with smaller confidence intervals than $\classical$. DSL is more variable, producing much less biased coefficients for some datasets (Multi-domain Sentiment, Bias in Biographies) but much more biased coefficients in others (Misinfo-general, Germeval18).

\begin{figure*}
    \centering
    \includegraphics[width=0.7\textwidth]{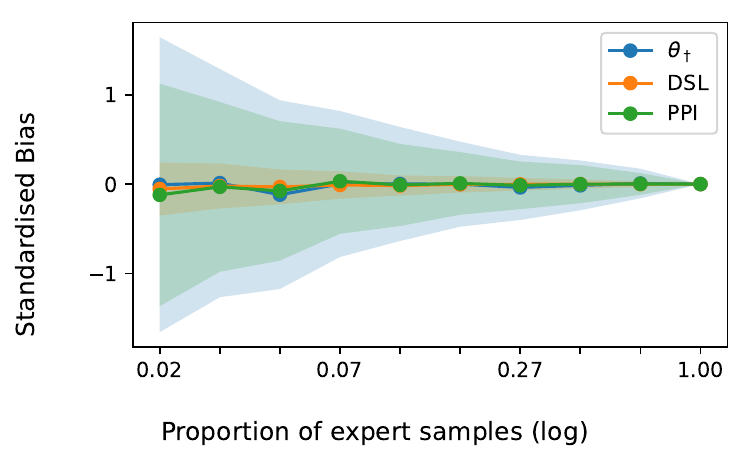}
    \caption{Performance of debiasing methods for Experiment 1 where highly correlated variables are removed, in terms of the standardized bias and aggregated over all datasets.}
    \label{fig:bias-all-no-collinear}
\end{figure*}

\begin{figure*}
    \centering
    \includegraphics[width=1\textwidth]{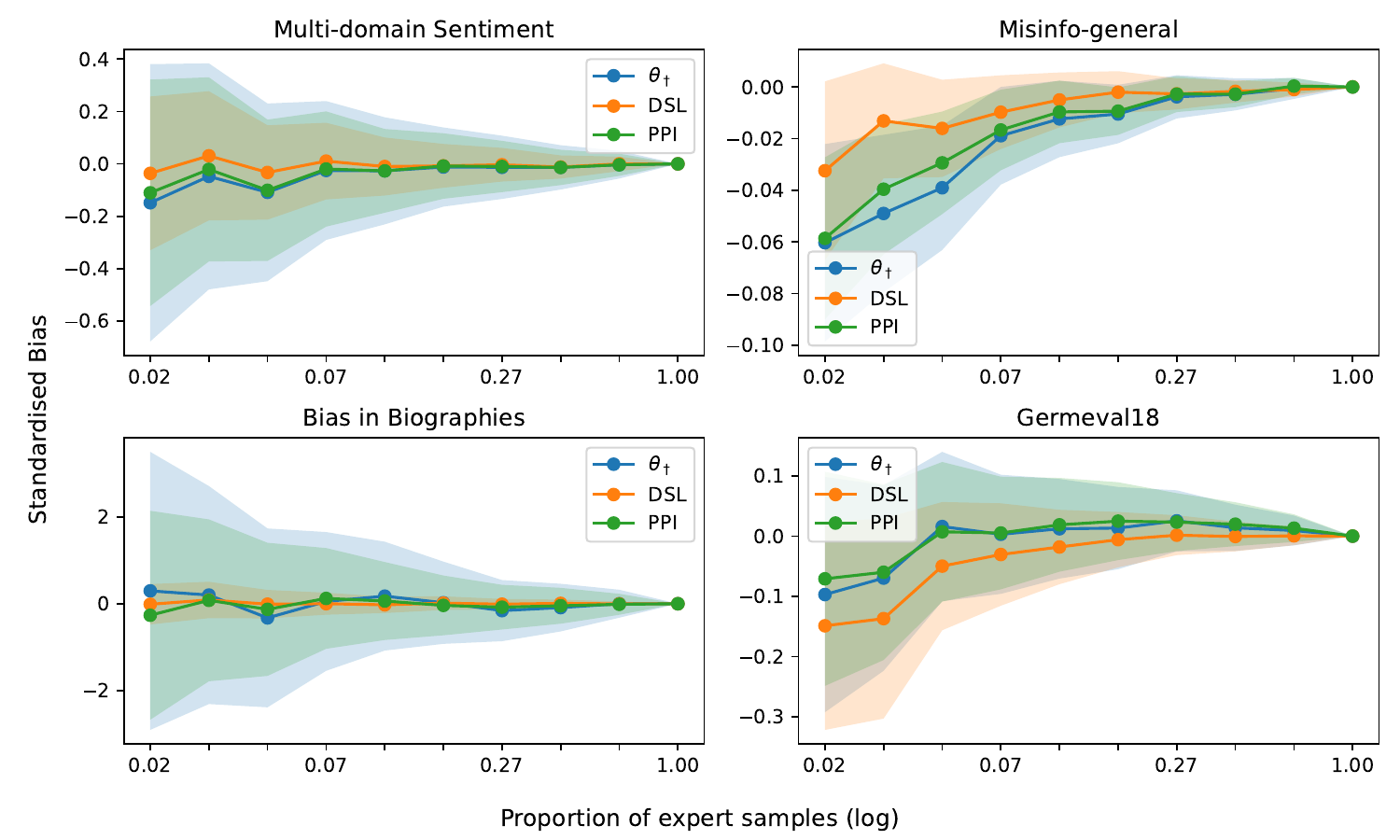}
    \caption{Performance of debiasing methods for Experiment 1, where highly correlated variables are removed, in terms of the standardized bias.}
    \label{fig:bias-datasets-no-collinear}
\end{figure*}

Figure \ref{fig:bias-all-no-collinear} and Figure \ref{fig:bias-datasets-no-collinear} show the results for the experiment from Appendix \ref{sec:removing-collinear-variables} where highly correlated features are removed. Compared to using all features, we notice a significant performance increase in all three methods. In particular, the coefficients produced by DSL are more stable.

\section{Comparison of DSL and PPI}

A comparison of DSL and PPI can be found in Table \ref{tab:ppi-dsl-contrast}.

\begin{table*}[ht]
\centering
\caption{Comparison of PPI \citep{ppi_original} and DSL\citep{dsl_original}  for debiasing ML predictions in downstream parameter estimation.}
\label{tab:ppi-dsl-contrast}
\begin{tabular}{l p{4cm} p{4cm} p{4cm}}
\toprule
\textbf{Method} & \textbf{Bias Correction Mechanism} & \textbf{Guarantees} & \textbf{Variance Components} \\
\midrule
PPI & First-order adjustment using estimating equation gradients (influence function) to offset systematic prediction errors. & Asymptotic validity holds irrespective of prediction model specification, provided large-sample coverage. & Aggregates prediction uncertainty and gradient-based correction variability; no design-specific terms. \\
\midrule
DSL & Post-estimation pseudo-outcome via doubly robust imputation, relying on specified selection probabilities. & Consistency if at least one of outcome regression or selection model is accurate; requires positive bounded probabilities. & Includes augmented regression variance plus inverse-probability weighting effects, which may be amplified under irregular sampling (e.g., covariate overlap, multi-collinearity). \\
\bottomrule
\end{tabular}
\end{table*}

\end{document}